%% file: CDAN.tex
\documentclass[letterpaper, 10 pt, conference]{ieeeconf}  

\makeatletter
\let\NAT@parse\undefined
\makeatother
\usepackage{amsmath}
\usepackage{amssymb}
\usepackage{acronym}
\usepackage{graphicx}
\usepackage{mdwtab}
\usepackage[numbers]{natbib}
\usepackage{hyperref}

\include{definitions}

\IEEEoverridecommandlockouts                              

\title{\LARGE \bf
Classifying Unordered Feature Sets \\with Convolutional Deep Averaging Networks
}
\author{Andrew Gardner$^{1}$, Jinko Kanno$^{1}$, Christian A. Duncan$^{2}$ and Rastko R. Selmic$^{3}$
\thanks{$^{1}$Andrew Gardner and Jinko Kanno are with the College of Engineering and Science, Louisiana Tech University,  Ruston, LA 71272, USA
        {\tt\small andrew.gardner1@ieee.org}, {\tt\small jkanno@latech.edu}}%
\thanks{$^{2}$Christian A. Duncan is with the Department of Engineering, Quinnipiac University,
        Hamden, CT 06518, USA
        {\tt\small christian.duncan@quinnipiac.edu}}%
\thanks{$^{3}$Rastko R. Selmic is with the Department of Electrical and Computer Engineering, Concordia University,
        Montreal, QC H4B 1R6, Canada
        {\tt\small rastko.selmic@concordia.ca}}
}

\begin{document}
\maketitle
\thispagestyle{empty}
\pagestyle{empty}
\input{acronyms}

\begin{abstract}
Unordered feature sets are a nonstandard data structure that traditional neural networks are incapable of addressing in a principled manner.
Providing a concatenation of features in an arbitrary order may lead to the learning of spurious patterns or biases that do not actually exist.
Another complication is introduced if the number of features varies between each set.
We propose convolutional deep averaging networks (CDANs) for classifying and learning representations of datasets whose instances comprise variable-size, unordered feature sets.
CDANs are efficient, permutation-invariant, and capable of accepting sets of arbitrary size.
We emphasize the importance of nonlinear feature embeddings for obtaining effective CDAN classifiers and illustrate their advantages in experiments versus linear embeddings and alternative permutation-invariant and -equivariant architectures.
\end{abstract}

\section{Introduction}
We propose \acp{cdan} for classifying and learning feature representations of datasets containing instances with unordered features, where each feature is considered a tuple composed of one or more values.
\acp{cdan} accept variable-size input and are invariant to permutations of the input's order.
In addition, as a side-effect of the training process, \acp{cdan} learn discriminative, nonlinear embeddings of individual input elements into a space of chosen dimensionality.
Contrary to their name, which is inspired by the work of Iyyer \etal\cite{imbgd-ducrsmtc}, \acp{cdan} could perhaps be more accurately termed convolutional deep pooling networks as we also consider the effects of functions other than averaging such as taking element-wise maximums or sums.

\subsection{Contributions}
We propose \acp{cdan} for classifying unordered feature sets.
We show that a \ac{cdan} with nonlinear embeddings is competitive with and perhaps even superior to \acp{rnn} and known permutation-invariant architectures for classifying instances containing variable-size sets of unordered features.
We also find that the type of pooling plays a significant role in determining the efficacy of the network with \texttt{sum}-pooling clearly outperforming \texttt{max}- and \texttt{average}-pooling.

\subsection{Related Research}
Sets, particularly those without an inherent ordering, comprise a class of data for which an obvious deep learning~\cite{lbh-dl} treatment is somewhat elusive.
A simple feed-forward neural network such as a \ac{mlp}~\cite{jmm-annat} is insufficient without enormous amounts of data and even more so if the sets are not of constant size.
In addition, \acp{rnn} are generally insufficient since the order of the elements may be unreliable or bias the network toward certain spurious or transient patterns.
Recently, the deep learning community has begun to explicitly consider architectures specifically made to address the unique challenges proposed by sets and other unusually structured data such as graphs~\cite{ccm-udhsg} and ordered sequences~\cite{vbk-omsss}.
These architectures usually work by exploiting or preserving symmetries in the data (see~\citet{gd-dsn} or~\cite{cw-gecn} for general frameworks).
In the remainder of the section, we focus on work more directly related to our own.

\citet{imbgd-ducrsmtc} proposed \acp{dan} for classifying text from an unordered list of words and showed that this rivaled more complex network architectures for the same task.
A \ac{dan} is essentially a traditional feed-forward neural network whose main distinguishing feature lies in the nature of its input: the element-wise average of word embeddings in a vector space.
\citeauthor{imbgd-ducrsmtc} did not consider learning word embeddings as part of the architecture, instead opting to use a set of predefined embeddings.
In addition, only averaging was considered as a means of aggregating the word embeddings.

\citet{hck-ldrsud} considered learning linear embeddings as part of the network architecture and summing instead of averaging the embeddings.
The resulting network was cast as an \ac{rnn} with identity weight matrices and served as a baseline against the article's primary architectures.
We show that linear embeddings are not sufficient for all tasks and indeed are unnecessary with certain pooling operations including averaging and summing.

\citet{rg-bwernnar} develop a neural bag-of-words model that is equivalent to a single-layer-embedding \ac{cdan} with average pooling.
Each dimension of the embedding is interpreted as the probability of a Gaussian-distributed visual word given the embedded element.
Consequently, the embedding is constrained by a softmax output.
\citeauthor{rg-bwernnar} do not appear to explicitly treat instances as sets rather than sequences, but their architecture is nevertheless permutation invariant.
A specialized layer representing a \ac{svm} with certain types of nonlinear kernels is incorporated after pooling.

Permutation equivariance is closely related to the concept of invariance.
Whereas invariance prescribes that the output of a function is unchanged when the input is permuted, equivariance indicates that the output (presumed to be a sequence or set of the same cardinality as the input) is permuted in the same manner as the input.
In other words, equivariance dictates that when a function $f:X^n \to Y^n$ is given $x \in X^n$ permuted by any $\pi \in \mathcal{S}_n$, where $\mathcal{S}_n$ is the symmetric group on $n$ symbols, then
\begin{equation}
f(\pi x) = \pi f(x).
\end{equation}
Note that invariance means that
\begin{equation}
f(\pi x) = f(x).
\end{equation}
\citet{rsp-dlspc} propose a computationally efficient permutation equivariant layer accomplished via a precise pattern of weight sharing.
The following equation computes the output $\vec{y}$ of a recommended version of this layer given an $n$ element $d$-dimensional input set represented as a matrix $\vec{x} \in \mathbb{R}^{n\times d}$,
\begin{equation}
\vec{y} = \vec{\sigma}(\vec{1}_n\transpose{\beta} + (\vec{x}-\vec{1}_n\transpose{\vec{x}_{\max}})\Gamma),
\label{eq:pmeqLayer}
\end{equation}
where $\vec{\sigma}$ is some nonlinear activation function, $\vec{x}_{\max} \in \mathbb{R}^d$ is a vector of the column-wise maximum values of $\vec{x}$, $\Gamma \in \mathbb{R}^{d \times m}$ is a weight matrix, $\beta \in \mathbb{R}^m$ is a bias, and $\vec{1}_n$ is a vector of $n$ ones.
\citet{gvwak-pennadp} also propose a permutation-equivariant layer for dynamics prediction but base their version on applying an arbitrary function to all pairwise combinations of input elements and averaging (pooling) the output, \ie given $n$ inputs $\vec{x}_i \in U$, $i\in [1,n]$ and a function $f:U \times U \to \mathbb{R}$, the $j$-th index of the output $\vec{y}$ is given by
\begin{equation}
y_j = \frac{1}{n}\sum_{i =1}^n f(\vec{x}_i, \vec{x}_j)
\label{eq:permutationalLayer}
\end{equation}
As noted by \citet{rsp-dlspc}, permutation invariance can be obtained from a permutation equivariant function by pooling over its output.

\citet{es-tns} propose a variational autoencoder~\cite{kw-aevb} for learning statistics of independent and identically distributed data. 
This work is perhaps the most similar to our own in that the proposed statistical network implicitly contains a \ac{cdan} as part of its structure.
The application of the implicit \ac{cdan} is distinguished from ours in that it is applied at the instance level rather than the feature level.
Whereas we are embedding individual features, \citeauthor{es-tns} embed instances.
In addition, \citeauthor{es-tns} appear to focus solely on average pooling.

\section{Convolutional Deep Averaging Networks}
Suppose we have a dataset $\mathbb{X}$ composed of $l$ subsets $X_i$, $i \in [1,l]$ of some set $U$ (theoretically, each $X_i$ may in fact be a multiset).
Let us assume $U \subset \mathbb{R}^d$ so that a given subset $X_i$ contains $n_i$ arbitrarily indexed vectors $\vec{x}^{(i)}_j$, $j \in [1,n_i]$.
Our objective is to design a neural network architecture capable of converting each of these variable-size subsets into a fixed-size representation that is useful for machine learning tasks such as classification.

One could certainly use an \ac{rnn} by treating each $X_i$ as a sequence.
However, if there is no inherent ordering to the elements, then an \ac{rnn} possesses some significant disadvantages.
The \ac{rnn} may learn or be biased towards spurious patterns that are a result of the chosen ordering scheme.
In addition, the removal of an element in the middle of the sequence could lead to unexpected results.

We reason that the ideal architecture for this problem is invariant to the order of the input, and we propose augmenting the \ac{dan} architecture by directly incorporating the embedding function $f: U \to \mathbb{R}^m$ into the structure of the network, where $m$ is the chosen size of the embedding.
We call the resulting architecture a \acl{cdan} due to its similarity to a \ac{cnn}, which will become apparent shortly.

\begin{figure}
\centering
\includegraphics[width = \linewidth]{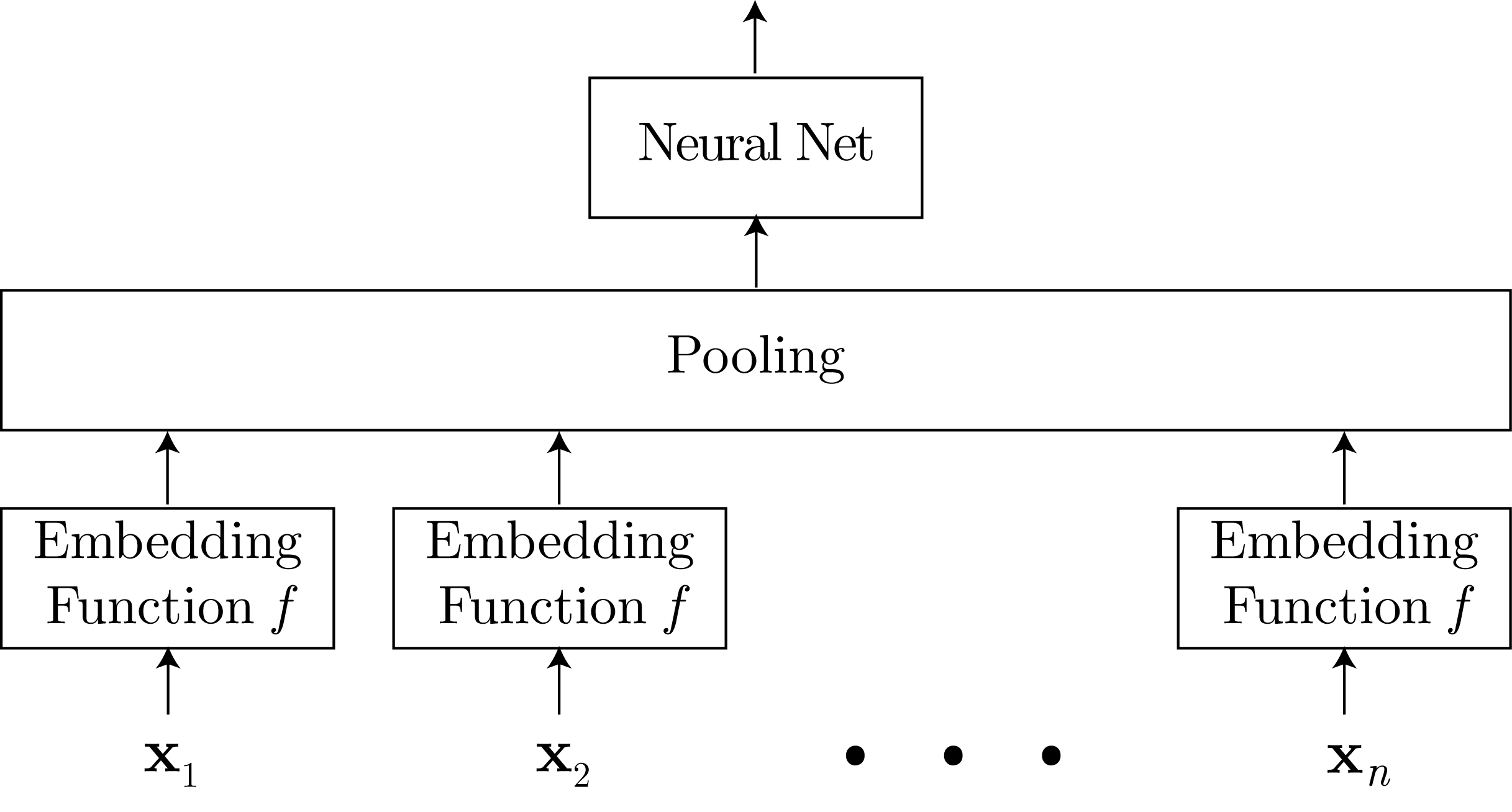}
\caption{An illustration of a generic \ac{cdan}. The inputs are arbitrarily indexed from 1 to $n$, where $n$ is the presumed cardinality of the input set. An embedding function $f$ is convolved with the input elements to produce a dynamically learned embedding in some potentially high-dimensional space.}
\label{fig:cdan}
\end{figure}

In theory, we place no restriction on the form that $f$ may take except that it be parameterized in a manner compatible with backpropagation-based training.
For the sake of simplicity, we assume that $f$ can be represented by an \ac{mlp}, although an \ac{rnn} is also conceivable if elements of $U$ are sequences or time series.
When given a set $X_i$, the embedding function is applied separately to each $\vec{x} \in X_i$.
One could informally interpret the embedding layer as a sort of convolution of $f$ with the elements of $X_i$.
The embeddings are then combined in a manner that does not depend on their order, \eg through a binary, commutative, and associative operator.
To borrow familiar language from \acp{cnn}, the embeddings are pooled.
Let $\rho: 2^{\mathbb{R}^m} \to \mathbb{R}^p$ denote the pooling function and note that usually $p=m$ as is the case for typical pooling operations such as summation.
A \ac{cdan} is then defined by the function $X \mapsto g(\rho(\mathcal{X}))$, where $\mathcal{X} = \{f(x) \suchthat x \in X\}$ and $g$ represents a neural network with arbitrary structure.
A \ac{cdan} with single-layer $f$ can be cast as a special type of \ac{cnn} by considering each set $X_i$ as an $n_i \times d$ image where $f$ is a bank of $m$ $1 \times d$ filters.
Alternatively, simply removing $\vec{1}_n\transpose{\vec{x}_{\max}}$ from~\refeq{pmeqLayer} yields an equivalent layer.
\acp{cdan} with \ac{mlp} embeddings may also be considered \acp{cnn} with multiple convolutional layers.
See~\reffig{cdan} for an illustration of the proposed architecture.

In an alternative interpretation of the embedding, we posit that $f$ effectively performs a type of bin or bucket sort of the set elements by allocating them to $m$ bins.
Each dimension of the embedding is thus associated with a certain region within the input space $U$.
Unlike an actual bucket sort, we do not require the bins to be disjoint.
By constraining the output of $f$ with a softmax function, however, one could produce a probability distribution over the bins.
This interpretation generalizes the neural bag-of-words model of Richard and Gall~\cite{rg-bwernnar} by allowing the distribution of each visual word to be learned rather than constrained to be Gaussian.
In a sense, such a network computes a probabilistic $k$-means with non-linear clusters.
Depending on the dimensionality of $U$, this interpretation provides us one way to visualize and examine the embedded feature space by plotting the activation of a bin or the distribution of the visual word in the input space.

The form of the embedding function plays a significant role in the performance of the network.
In the following subsections, we show that nonlinear embeddings are generally preferable to linear.

\subsection{Disadvantage of Linear Embeddings}
Consider a linear embedding $f_\text{lin}: U \to \mathbb{R}^m$ defined by 
\begin{equation}
f_\text{lin}(x) = \transpose{W}\vec{x}+\vec{b},
\end{equation}
where $W$ is an $d \times m$ weight matrix and $\vec{b} \in \mathbb{R}^m$ is a bias vector.
Assume that the pooling layer consists of an \texttt{average} operation.
The output of the pooling layer and input to the deep portion of the network given $X_i$ is then 
\begin{equation}
\begin{split}
\frac{1}{n_i}\sum_{j=1}^{n_i}f_\text{lin}\left(\vec{x}^{(i)}_j\right) &= \frac{1}{n_i}\sum_{j=1}^{n_i} \left[\transpose{W}\vec{x}^{(i)}_j + \vec{b}\right]\\
 &= \transpose{W}\sum_{j=1}^{n_i}\frac{\vec{x}^{(i)}_j}{n_i} + \vec{b} \\
&= f_\text{lin}\left(\frac{1}{n_i}\sum_{j=1}^{n_i}\vec{x}^{(i)}_j\right).
\end{split}
\end{equation}
We see that we could have simply pooled the input elements directly.
In addition, if $V$ and $\vec{c}$ are the weights and bias of the first post-pooling layer, then $f_\text{lin}$ could be merged into the layer by substituting $V$ and $\vec{c}$ with $WV$ and $\transpose{V}\vec{b}+\vec{c}$.
In other words, the linear embedding is computationally unnecessary and can be eliminated.
A similar conclusion may be reached if \texttt{sum}-pooling is used instead (or any linear operation).
\texttt{Max}-pooling is an exception as it introduces a nonlinearity.
However, \texttt{max}-pooling with linear embeddings still has potential issues with ambiguity.

\subsection{Nonlinear Embeddings Mitigate Ambiguity}
Based on the previous subsection's result, one may consider simply skipping a learned embedding and working directly with the input points as the plain \ac{dan} of \citet{imbgd-ducrsmtc} suggests.
In general, though, this course of action may be unwise.
In particular, suppose there are two sets $X_i$, $X_j$ such that
\begin{equation}
\sum_{\vec{x}_i \in X_i} \vec{x}_i = \sum_{\vec{x}_j \in X_j} \vec{x}_j.
\end{equation}
One could even construct a situation wherein both sets also have the same element-wise maximums by choosing $X_i$ and $X_j$  to have the same convex hull.
In such an event, $X_i$ and $X_j$ are indistinguishable under linear embeddings with \texttt{max}-pooling since the maximum (and minimum) of a linear function will always lie on the boundary (\ie vertices) of a convex set.
Regardless of the cause of the ambiguity, the consequence is that instances with potentially significant differences are functionally identical from the network's perspective.
The primary issue, though, is the fact that these ambiguities are not caused by particularly exotic circumstances.

\begin{figure}
\centering
\includegraphics[width=\linewidth]{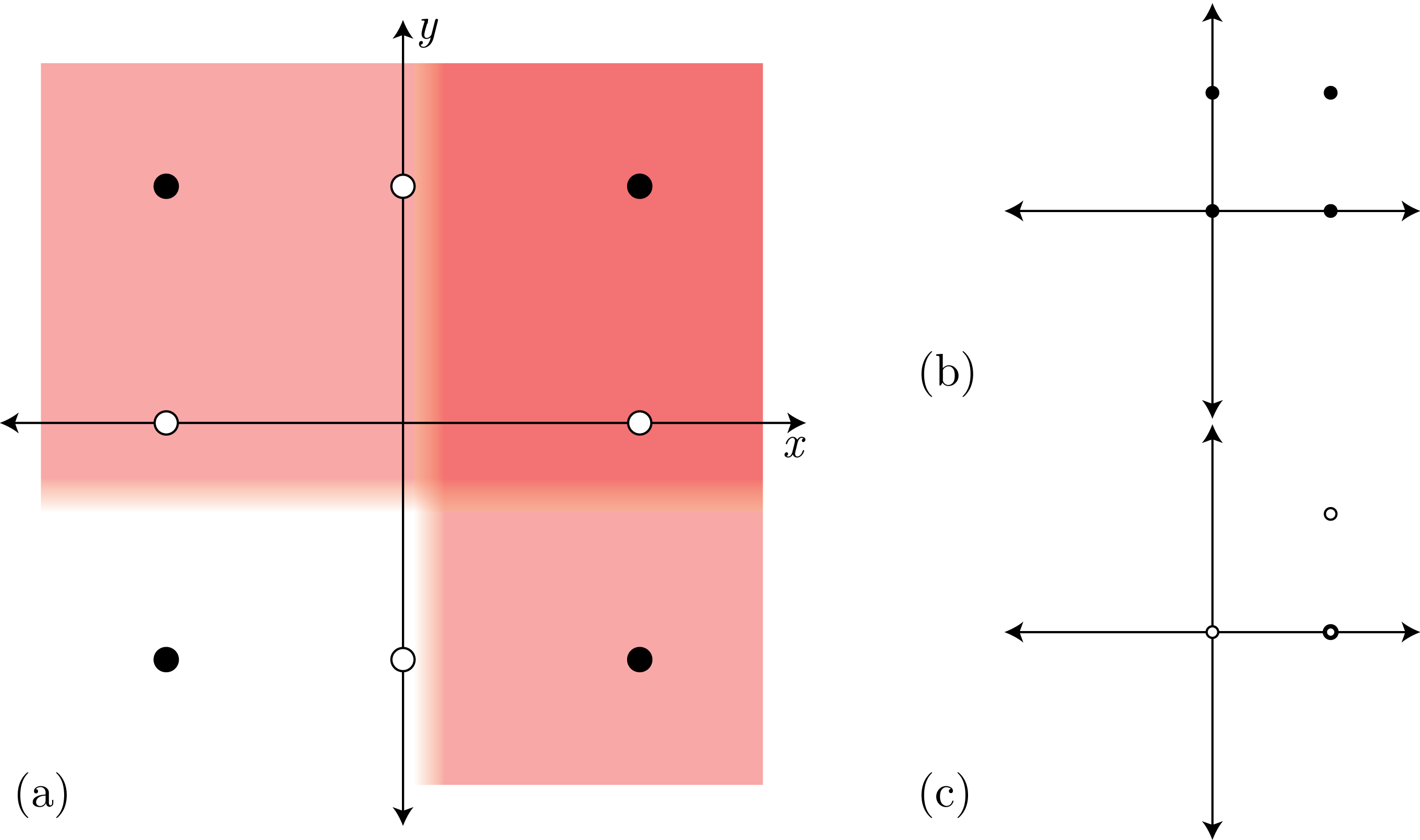}
\caption{An example of simultaneous \texttt{sum}-, \texttt{average}-, and \texttt{max}-pooling ambiguity and its partial resolution via a nonlinear embedding. (a)  The set of black points and the set of white points shown have the same coordinate-wise sums and maximums. The shading shows the activation of two sigmoidal functions that can be used to construct nonlinear 2D embeddings (b and c) that distinguish the two sets under \texttt{sum}- and \texttt{average}-pooling. (b) The embedding of the black points. (c) The embedding of the white points. Note that two points share nearly the same embedding.}
\label{fig:ambiguityExample}
\end{figure}

A nonlinear embedding allows the network to learn functions that can differentiate sets that are ambiguous under linear pooling. 
Note that ambiguity is still possible with a nonlinear embedding.
However, since the embedding is learned to satisfy some objective, one can expect these ambiguities to either be benign or to indicate some inherent similarity between the ambiguous instances.
For example, consider the sets of black and white points in~\reffig{ambiguityExample}(a) that are ambiguous under \texttt{sum}- and \texttt{max}-pooling.
Using a pair of sigmoidal activation functions each defined by
\begin{equation}
\sigma(a) = \frac{1}{1+e^{-a}},
\end{equation}
with inputs $a_1 = x \pm \epsilon_1$, $a_2 = y \pm \epsilon_2$, where $\epsilon_1$, $\epsilon_2$ are each small and positive, we can compute nonlinear embeddings that are unambiguous under \texttt{sum}-pooling.

The nonlinear embedding of the entire set is the key point; linear point-wise embeddings followed by \texttt{max}-pooling may be sufficient when equivalent convex hulls are rare.
However, we hypothesize that nonlinear embeddings are inherently more powerful and thus more useful since they have greater representational capacity.

\section{Experiments}
We conduct experiments to evaluate the performance of \acp{cdan} against alternative architectures as well as examine the effects of different pooling operations.
Our experiments focus especially on variable-size sets, which do not seem to have many existing results in the literature.
All models were implemented and tested using the Keras~\cite{keras} deep learning framework with the Theano~\cite{theano-short} backend.

\subsection{Posture Recognition from Point Sets}
A motion capture dataset of hand postures provides the primary basis for our evaluation.\footnote{The dataset along with further documentation is available at \url{http://www.latech.edu/~jkanno/collaborative.htm}.}
The dataset consists of variable-size point sets representing five hand postures captured from 12 users.
The size of each point set ranges from 3 to 12, although it should be noted that only 11 markers were physically present. 
Each point set shares the same coordinate system, so no rotations or translations should be required to process the data.
Regardless, we center each point set to have zero mean in each dimension.
The goal is to classify each point set as one of the five postures.

In order to make the problem more challenging, we do a leave-one-user-out evaluation where all but one user contribute to the training and validation sets and the test set is drawn exclusively from the left-out user.
Each user is iteratively left out, and the resulting test accuracies are averaged to obtain a reasonable evaluation of the tested classifier's generalization error.
Training, validation, and test sets are disjoint and each consist of 75 uniformly randomly selected instances per class per user without replacement.
This process is repeated five times in order to obtain some measure of confidence in the results.

\subsection{Model Specification}
We compare a variety of \ac{cdan} architectures for this task, including linear embeddings with \texttt{max}-pooling, linear embedding with \texttt{sum}-pooling (\ie no embedding), and nonlinear embeddings with \texttt{average}-, \texttt{sum}-, and \texttt{max}-pooling.
These models are compared against an \ac{rnn} with \acp{gru}~\cite{cgcb-eegrnnsm} as well as an experimental variant of the \ac{cdan} architecture with recurrent connections between the embeddings, which we call a \ac{rdan}.
In an \ac{rdan}, we trade permutation invariance and independent embeddings for increased functional capacity.
Note that an \ac{rdan} is effectively just an \ac{rnn} that is pooled over the entire time axis.
Finally, we implement the permutation-equivariant layers of \citet{gvwak-pennadp} (defined by~\refeq{permutationalLayer}) and \citet{rsp-dlspc} (defined by~\refeq{pmeqLayer}) for an external, contemporaneous comparison.
One or more permutational layers enable one to obtain dependent nonlinear embeddings that are permutation invariant (after pooling) as opposed to the \acp{rnn}.
From this point forth, we refer to the permutational layer of \citeauthor{gvwak-pennadp} as a pairwise layer due to its structure and to distinguish it from the permutation-equivariant layer of \citeauthor{rsp-dlspc}. 
We refer to the respective types of model as \acp{pcdan} and \acp{pdan} for brevity.
Despite the fact that the nonlinear embeddings of a \ac{cdan} are not always technically convolutions, we will sometimes refer to them as convolutional layers when compared against the recurrent, pairwise, and permutational layers of competing architectures.

Given the incredibly diverse array of architectural and training options available in the literature, we tried to make our architectures as uniform as possible in order to enable fair comparison.
Since the recurrent architectures depend on the order of the input, points in each set were lexicographically sorted by their $x$-, $y$- and $z$-coordinates.
Gaussian noise with a standard deviation of 20 (millimeters, which is the scale of the input) was applied to the input as a form of regularization for each network.
Dropout~\cite{shkss-dswpnno} of 10\% was applied to the hidden layers of each network, and $l_2$ regularization with a magnitude of 0.001 was applied to the weights of each layer.
We did not apply dropout to the input, but we did adopt the simultaneous dropout suggested by \citet{rsp-dlspc} for the \acp{pdan}, which consists of dropping a feature simultaneously in all elements of an input set rather than independently.
A default embedding size of 11 was chosen for computational expedience as well as to let each embedded dimension hypothetically represent one of the physical markers.
Two special (\ie convolutional, recurrent, \etc) layers with 11 neurons each were used in each architecture.
Each tested recurrent network was bidirectional~\cite{sp-brnn} with the forward and backward \ac{rnn} outputs concatenated at each timestep (\ie 22 dimensional output).
To clarify, the final timesteps in each direction were concatenated in the case of the plain \ac{rnn} without pooling.
Except in the case of linear embeddings, maxout activations~\cite{gwfmcb-mn} with 2 pieces were used in each layer except for the network output, which incorporated a softmax activation.
Each model used the same post-embedding architecture, which consisted of one 11-neuron layer with a residual connection~\cite{hzrs-drlir} followed by the 5-neuron (one per class) softmax layer.

\acp{cdan} offer significant computational and practical advantages over the other architectures that arise primarily from the fact that the embeddings are independent.
Unlike an \ac{rnn} or \ac{rdan}, the embeddings can be computed in parallel rather than sequentially.
In addition, only $O(n)$ embedding function evaluations are required as opposed to $O(n^2)$ function evaluations for a \ac{pcdan}.
The fact that the embeddings are independent also enables their re-use in intersecting sets whereas recurrent or permutational architectures must re-evaluate each point.
\acp{pdan} are of similar complexity, although they lack any advantages derived from independent embeddings.
For these reasons we were able to experiment with \acp{cdan} and \acp{pdan} with embedding sizes that are an order of magnitude higher than the other models (100 to be precise) yet still require less computation.

The RMSProp~\cite{th-rmsprop} implementation provided by Keras~\cite{keras} was used with a learning rate of 0.001 and minibatches of size 64.
Training was terminated for a model if the validation loss did not improve after 40 epochs.

\subsection{Results and Discussion}
Results are presented in~\reftab{postureAccs}, where we can see that the highest average accuracy was achieved by a \ac{cdan} with \texttt{sum}-pooling and 100-dimensional nonlinear embeddings.
We also immediately notice a significant difference between types of pooling for permutation invariant architectures.
\acp{rdan}, on the other hand, appear to be robust to changes in the pooling mode and just as effective if not marginally better than the \ac{rnn}.
\begin{table}[h!]
\centering
\caption{Average accuracies and standard deviations over five leave-one-user-out evaluations.}
\label{tab:postureAccs}
\begin{tabular}{|l|| c|c| c| c|}
\hlx{hv}
\textbf{Type} & \textbf{Embedding} & \textbf{Embedding} & \textbf{Pooling} & \textbf{Accuracy}\\
\textbf{} & \textbf{} & \textbf{Size} & \textbf{} & \\
\hlx{vhv}
\acs{cdan} & None & N/A & \texttt{sum} & $\underset{\pm 0.10}{20.05}$\\
\acs{cdan} & Linear & 11 & \texttt{max} &  $\underset{\pm 4.30}{70.00}$\\
\acs{cdan} & Linear & 100 & \texttt{max} &  $\underset{\pm 0.96}{75.94}$\\
\acs{cdan} & Nonlinear & 11 & \texttt{average} & $\underset{\pm 1.00}{71.61}$\\
\acs{cdan} & Nonlinear & 11 & \texttt{max} & $\underset{\pm 1.85}{65.44}$\\
\acs{cdan} & Nonlinear & 11 & \texttt{sum} & $\underset{\pm 1.37}{89.14}$\\
\acs{cdan} & Nonlinear & 100 & \texttt{average} & $\underset{\pm 1.81}{77.00}$\\
\acs{cdan} & Nonlinear & 100 & \texttt{max} & $\underset{\pm 1.60}{75.81}$\\
\acs{cdan} & Nonlinear & 100 & \texttt{sum} & $\underset{\pm 1.72}{92.24}$\\
\hlx{vhv}
\acs{rdan} & Recurrent & 11 & \texttt{average} & $\underset{\pm 1.15}{90.25}$\\
\acs{rdan} & Recurrent & 11 & \texttt{max} & $\underset{\pm 0.98}{90.31}$\\
\acs{rdan} & Recurrent & 11 & \texttt{sum} & $\underset{\pm 1.25}{90.38}$\\
\acs{rnn} & Recurrent & 11 & N/A & $\underset{\pm 1.59}{89.28}$\\
\hlx{vhv}
\acs{pcdan} & Pairwise & 11 & \texttt{average} & $\underset{\pm 1.91}{85.86}$\\
\acs{pcdan} & Pairwise & 11 & \texttt{max} & $\underset{\pm 0.59}{81.94}$\\
\acs{pcdan} & Pairwise & 11 & \texttt{sum} & $\underset{\pm 1.45}{89.67}$\\
\hlx{vhv}
\acs{pdan} & Permutational & 11 & \texttt{average} & $\underset{\pm 1.41}{70.64}$\\
\acs{pdan} & Permutational & 11 & \texttt{max} & $\underset{\pm 0.59}{61.31}$\\
\acs{pdan} & Permutational & 11 & \texttt{sum} & $\underset{\pm 2.17}{76.59}$\\
\acs{pdan} & Permutational & 100 & \texttt{average} & $\underset{\pm 2.86}{80.84}$\\
\acs{pdan} & Permutational & 100 & \texttt{max} & $\underset{\pm 0.67}{77.65}$\\
\acs{pdan} & Permutational & 100 & \texttt{sum} & $\underset{\pm 1.53}{87.31}$\\
\hlx{vh}
\end{tabular}
\end{table}

In general, we note that the highest accuracies achieved for each type are clustered around 90\% accuracy.
Our results do not provide enough confidence to say that the best-performing models are significantly different (in a statistical sense) than one another, but they do suggest a potential advantage to certain \acp{cdan} and disadvantage to \ac{pdan}.
The \acp{pdan} slightly inferior performance may be explained by the fact that their permutation-equivariant layers are slightly more constrained than the competition.
Furthermore, we tested only a portion of the possible architectures proposed by the framework of \citet{rsp-dlspc}.
Regardless of whether the best \ac{cdan} does achieve significantly higher accuracy than the competition, the computational advantages of a \ac{cdan} over \acp{rnn} and \acp{pcdan} certainly warrants their utility.
In particular, reducing the embedding size to 11 renders a significantly more efficient classifier (with relatively few parameters) with only a marginal drop in accuracy.

We can hypothesize potential reasons for the pattern of results induced by different pooling modes.
Note first that the difference between \texttt{average}- and \texttt{sum}-pooling must arise from the fact that the input sets are not of constant size.
If the size was constant, then both pooling modes would be the same but for a constant factor. 
A potential cause for their difference here may thus arise from the fact that \texttt{average}-pooling effectively removes information (the implicitly encoded size of the set) and introduces ambiguity between certain set embeddings.
On the other hand, \texttt{max}-pooling's relatively poor performance may be partially due to the choice of the maxout activation function.
We noted in some exploratory trials that its accuracy significantly improved when paired with \ac{relu} activations.
Some theoretical basis for \texttt{sum}-pooling's apparent advantage may be given by a probabilistic interpretation. 
Though embeddings were not constrained by a softmax output, we may interpret them as the logarithm of unscaled posterior probabilities as indicated by a neural bag-of-words model~\cite{rg-bwernnar}.
The sum of the embeddings then gives the log-likelihood (shifted by some amount) for the parameters associated with each visual word given the point set.

We also show that nonlinear embeddings can yield significant gains over linear or identity (\ie no) embeddings. 
Indeed, for this problem the identity embedding yields a network no better than guessing.
The linear embedding with \texttt{max}-pooling, on the other hand, is competitive with its counterpart nonlinear embedding.
However, whereas the nonlinear embedding could potentially be improved by adding more layers or changing its activation functions, the linear embedding is already exhausting its functional capacity.

\section{Conclusion}
We introduced the \ac{cdan}, a class of neural networks designed for classifying instances containing unordered, variable-size feature sets.
The proposed architecture works by directly incorporating a function into the network's structure that embeds the features in a high-dimensional space and pooling the subsequent embeddings.
As the name implies, an equivalence can be drawn between the convolution operation in \acp{cnn} and the application of the embedding function.
Experiments show that in terms of accuracy, \acp{cdan} are competitive with competing recurrent and permutation-equivariant architectures.
\acp{cdan} are also computationally efficient compared to alternative architectures, favoring parallel implementations and re-use of prior results since feature embeddings are set-invariant.
In addition, the learned feature embeddings are a useful by-product that can potentially solve related problems such as nonlinear clustering. 

Future work may include further exploration of \ac{cdan} properties and optimal architectures when applied to other problems or datasets.
One should note that networks incorporating convolutional, recurrent, or permutation-equivariant layers need not be mutually exclusive.
Architectures that perform a convolutional embedding prior to a permutation-equivariant layer (or vice-versa) may be worth exploring and could be capable of achieving results superior to either method when used alone.


\bibliographystyle{abbrvnat}
\bibliography{references}
\end{document}

%% file: definitions.tex
\newif\ifIEEE
\IEEEtrue

\newcommand{\refeq}[1]{(\ref{eq:#1})}

\newcommand{\reftab}[1]{Table~\ref{tab:#1}}

\newcommand{\reffig}[1]{Figure~\ref{fig:#1}}

\renewcommand{\vec}[1]{\boldsymbol{\mathbf{#1}}}

\newcommand{\transpose}[1]{{{}#1}_{\phantom{0}}^{\intercal}}

\newcommand{\expected}[2]{\ifthenelse{\isempty{#2}}{\mathbb{E}}{\mathbb{E}_{#2}}\left[{#1}\right]}

\newcommand{\suchthat}{\;\ifnum\currentgrouptype=16 \middle\fi|\;}

\newcommand{\approptoinn}[2]{\mathrel{\vcenter{
  \offinterlineskip\halign{\hfil$##$\cr
    #1\propto\cr\noalign{\kern2pt}#1\sim\cr\noalign{\kern-2pt}}}}}

\newcommand\etal{\textit{et~al.\@~}}
\newcommand\etc{etc.\@}
\ifIEEE
\newcommand\ie{i.e.\ }
\newcommand\eg{e.g.\ }

\else
\newcommand\ie{that is, }
\newcommand\eg{for example, }

\fi

\newcommand\EMDHat[1][]{\ifthenelse{\isempty{#1}}{$\widehat{\text{EMD}}$}{$\widehat{\text{EMD}}_{#1}$}}
\newcommand\EMDNot[1][]{\ifthenelse{\isempty{#1}}{$\overline{\text{EMD}}$}{$\overline{\text{EMD}}_{#1}$}}

%% file: acronyms.tex
\acrodef{ekf}[EKF]{extended Kalman filter}
\acrodef{ann}[ANN]{artificial neural network}
\acrodef{mlp}[MLP]{multi-layer perceptron}
\acrodef{rnn}[RNN]{recurrent neural network}
\acrodef{cnn}[CNN]{convolutional neural network}
\acrodef{gru}[GRU]{gated recurrent unit}
\acrodef{kkt}[KKT]{Karush-Kuhn-Tucker}
\acrodef{relu}[ReLU]{rectified linear unit}
\acrodef{gmm}[GMM]{Gaussian mixture model}
\acrodef{ta}[TA]{Technology Area}
\acrodef{hmm}[HMM]{hidden Markov model}
\acrodef{csv}[CSV]{comma separated value}
\acrodef{pca}[PCA]{principal component analysis}

\acrodef{afem}[AFEM]{\textit{a fortiori} expectation-maximization}
\acrodef{em}[EM]{expectation-maximization}
\acrodef{jpda}[JPDA]{joint probabilistic data association}
\acrodef{js}[JS]{Jensen-Shannon}
\acrodef{kl}[KL]{Kullback-Leibler}
\acrodef{mcmc}[MCMC]{Markov chain Monte Carlo}
\acrodef{phd}[PHD]{probability hypothesis density}

\acrodef{emd}[EMD]{earth mover's distance}
\acrodef{vbow}[VBOW]{Visual Bag-of-Words}
\acrodef{svm}[SVM]{support vector machine}
\acrodef{ksvm}[KSVM]{Krein support vector machine}
\acrodef{pd}[PD]{positive definite}
\acrodef{nd}[ND]{negative definite}
\acrodef{cpd}[CPD]{conditionally positive definite}
\acrodef{cnd}[CND]{conditionally negative definite}
\acrodef{rbf}[RBF]{radial basis function}
\acrodef{pde}[PDE]{partial differential equation}
\acrodef{emi}[EMI]{earth mover's intersection}
\acrodef{emjd}[EMJD]{earth mover's Jaccard distance}
\acrodef{emji}[EMJI]{earth mover's Jaccard index}
\acrodef{ospa}[OSPA]{optimal subpattern assignment}

\acrodef{ber}[BER]{balanced error rate}
\acrodef{knn}[$k$-NN]{$k$-nearest neighbor}
\acrodef{fs}[FS]{feature selection}

\acrodef{dan}[DAN]{deep averaging network}
\acrodef{cdan}[CDAN]{convolutional deep averaging network}
\acrodef{rdan}[RDAN]{recurrent deep averaging network}
\acrodef{pcdan}[PCDAN]{pairwise convolutional deep averaging network}
\acrodef{pdan}[PDAN]{permutational deep averaging network}